\documentclass[12pt]{article}
\usepackage{fullpage}
\usepackage{amssymb, amsthm, amsmath}
\usepackage{setspace}
\usepackage{bm}
\usepackage{graphicx}
\usepackage[authoryear]{natbib}
\usepackage{verbatim}
\usepackage{lineno}
\usepackage{times}
\usepackage{caption}
\usepackage{subcaption}
\usepackage{epstopdf}
\usepackage{bbm}
\usepackage{lscape}
\usepackage{multirow}
\usepackage[bottom]{footmisc}
\usepackage[normalem]{ulem}
\usepackage{color}
\usepackage{xcolor}
\usepackage{censor}
\usepackage{hyperref}
\usepackage{makecell}
\usepackage{appendix}
\usepackage{etoolbox}
\patchcmd{\appendices}{\quad}{: }{}{}

\usepackage{appendix}
\usepackage{titlesec}
 \usepackage{etoolbox, chngcntr}
\AtBeginEnvironment{appendices}{%
 \titleformat{\section}{\bfseries\Large}{\appendixname~\thesection:}{0.5em}{}%
 \titleformat{\subsection}{\bfseries\large}{\thesubsection}{0.5em}{}%
\counterwithin{equation}{section}
}

\newcommand\hl[1]{%
  \bgroup
  \hskip0pt\color{blue!100!black}%
  #1%
  \egroup
}

\newcommand{\q}{q}

\newcommand{\Y}{Y}
\newcommand{\y}{y}

\newcommand{\Z}{Z}
\newcommand{\z}{z}
\newcommand{\ProjectTwoZi}{\Z_\obs}
\newcommand{\ProjectTwozi}{\z_\obs}
\newcommand{\zilstar}{\z^*_{\obs\control}}
\newcommand{\zionestar}{\z_{\obs1}^*}
\newcommand{\zl}{\z_\testingresponsevalind}
\newcommand{\zik}{\z_{\obs\control}}
\newcommand{\dz}{d\z}

\newcommand{\parquantbmeanparam}{\bm{\beta}}
\newcommand{\parquantsdparam}{\sigma}

\newcommand{\integralstdresponse}{u}

\newcommand{\qziXitheta}{\q(\ProjectTwozi,\bXi;\paramvec)}
\newcommand{\qzikstarXltheta}{\q(\z^{*}_{\obs\control};\bXi,\paramvec)}
\newcommand{\qzXtheta}{\q(\z,\bX;\paramvec)}
\newcommand{\qzjXthetahat}{\q(\z_j,\bX;\estparamvec)}
\newcommand{\qziX}{\q(\ProjectTwozi,\bX)}
\newcommand{\qzX}{\q(\z,\bX)}
\newcommand{\quX}{\q(\integralstdresponse,\bX)}
\newcommand{\qzXitheta}{\q(\z,\bXi;\paramvec)}
\newcommand{\qzikXitheta}{\q(\zik,\bXi;\paramvec)}

\newcommand{\conddistfxnyX}{h(\y|\bX)}
\newcommand{\conddistfxnziX}{f(\ProjectTwozi|\bX)}
\newcommand{\conddistfxnzlXestparamvec}{f(\zl|\bX;\estparamvec)}
\newcommand{\conddistfxnzX}{f(\z|\bX)}

\newcommand{\polynomialmodelparamtwo}{\upsilon}
\newcommand{\polynomialmodelparam}{\xi}
\newcommand{\polynomialmodelparamvec}{\bm{\polynomialmodelparam}}

\newcommand{\maxquadraticterm}{o}
\newcommand{\maxquadraticterms}{O}

\newcommand{\covtwo}{k}

\newcommand{\params}{m}

\newcommand{\polynomialpower}{b}
\newcommand{\polynomialpowers}{B}

\newcommand{\testingresponsevalind}{l} 
\newcommand{\testingresponsevalinds}{L}

\newcommand{\MVN}{\text{MVN}}

\newcommand{\intensityfxn}{\lambda}
\newcommand{\IntensityFxn}{\Lambda}

\newcommand{\paramraw}{\theta}
\newcommand{\paramvec}{\bm{\paramraw}}
\newcommand{\estparamvec}{\hat{\paramvec}}

\newcommand{\X}{X}
\newcommand{\bX}{\bm{\X}}
\newcommand{\bXi}{\bX_\obs}

\newcommand{\inputlayernode}{I}
\newcommand{\hiddenlayernode}{H}

\newcommand{\inputlayercoef}{\beta}
\newcommand{\hiddenlayercoef}{\gamma}
\newcommand{\outputlayercoef}{\delta}

\newcommand{\outputlayercoefi}{\outputlayercoef_{\hiddenlayernodeindex}}
\newcommand{\hiddenlayercoefi}{\hiddenlayercoef_{\hiddenlayernodeindex\inputlayernodeindex}}
\newcommand{\inputlayercoefi}{\inputlayercoef_{\cov\inputlayernodeindex}}

\newcommand{\ridgepenalty}{\omega}

\newcommand{\activationfunction}{f_A}

\newcommand{\cov}{j}
\newcommand{\covs}{p}

\newcommand{\obs}{i}
\newcommand{\obss}{n}

\newcommand{\hiddenlayernodeindex}{t}
\newcommand{\hiddenlayernodeindexes}{T}

\newcommand{\inputlayernodeindex}{r}
\newcommand{\inputlayernodeindexes}{R}

\newcommand{\control}{k}
\newcommand{\controls}{K}

\begin{document}

\begin{center}
	{\Large Nonparametric conditional density estimation in a deep learning framework for short-term forecasting}\\\vspace{6pt}
	{\large David B. Huberman \footnote{Department of Statistics, North Carolina State University; Campus Box 8203; Raleigh, NC 27695, United States; dbhuberm@ncsu.edu}, Brian J. Reich\textsuperscript{1}, \setcounter{footnote}{1} and Howard D. Bondell \footnote{\setstretch{.5} School of Mathematics and Statistics, University of Melbourne, Peter Hall Building, VIC 3122, Australia}}\\
	{\large North Carolina State University}\\
	\today
\end{center}

\begin{abstract}\begin{singlespace}

Short-term forecasting is an important tool in understanding environmental processes. In this paper, we incorporate machine learning algorithms into a conditional distribution estimator for the purposes of forecasting tropical cyclone intensity. Many machine learning techniques give a single-point prediction of the conditional distribution of the target variable, which does not give a full accounting of the prediction variability. Conditional distribution estimation can provide extra insight on predicted response behavior, which could influence decision-making and policy. We propose a technique that simultaneously estimates the entire conditional distribution and flexibly allows for machine learning techniques to be incorporated. A smooth model is fit over both the target variable and covariates, and a logistic transformation is applied on the model output layer to produce an expression of the conditional density function. We provide two examples of machine learning models that can be used, polynomial regression and deep learning models. To achieve computational efficiency we propose a case-control sampling approximation to the conditional distribution. A simulation study for four different data distributions highlights the effectiveness of our method compared to other machine learning-based conditional distribution estimation techniques. We then demonstrate the utility of our approach for forecasting purposes using tropical cyclone data from the Atlantic Seaboard. This paper gives a proof of concept for the promise of our method, further computational developments can fully unlock its insights in more complex forecasting and other applications.

\end{singlespace}

\begin{singlespace}
\end{singlespace}

\end{abstract}
	\newpage	

\section{Introduction}

Short-term forecasting of environmental processes has many applications including solar and wind power generation, ambient air pollution, and extreme weather events.  In this paper, we combine numerical model output with statistical methods to forecast hurricane wind intensity.  Rather than providing a single value as the point prediction, we model the entire uncertainty distribution of the response given the numerical model forecast.  This conditional distribution regression provides a comprehensive assessment of uncertainty, including the forecast distribution's spread, skewness and tail probabilities.

To provide a flexible prediction model, we incorporate supervised machine learning methods, which have become a popular tool for statistical analysis in the last few decades. Methods such as random forest regression, neural networks, and linear regression can be employed using state-of-the-art statistical software to clarify complicated relationships between covariates and target variables. Generally, machine learning predictive modeling has been developed for making point predictions such as the conditional mean or median. Accompanying prediction interval techniques provide uncertainty quantification. This differs from conditional density estimation, a technique which estimates the full distribution of the target variable given the covariates. In some applications, conditional density estimation is preferred. For instance, an estimate of a tropical cyclone's maximum wind speed conditional on the sea surface temperature can provide information not available from a conditional mean estimate. A certain sea surface temperature might result in a strongly positively skewed maximum wind speed distribution, giving a better idea of the worst case scenario under these conditions.

Various approaches have been developed to estimate the distribution of the target variable conditional on the covariates. One technique is to estimate the joint distribution of the target variable and covariates as well as the joint distribution of the covariates and divide the former by the latter. Kernel density estimation of these two densities is a common approach, first proposed by \citet{rosenblatt1969}. \citet{hyndman1996} modify the standard kernel density estimator to obtain a smoother with better bias properties. \citet{hall1999} propose to use an adjusted Nadaraya-Watson estimator for the kernel estimation. These methods suffer from intractability when the covariate dimension increases. The proposed remedies for this issue have been modifications to reduce the covariate space or to develop a density estimator for high-dimensional data \citep{hall2004, hall2005, fan2009}. 

Bayesian nonparametric mixture modeling is another common conditional density estimation approach.
Finite mixture models (FMMs) are a subset of mixture modeling techniques which consider the conditional target distribution to be a mixture of several parametric (often Gaussian) distributions \citep{escobar1995, gilardi2002, song2004, rojas2005, fahey2007}. Covariate effects can be introduced in either the mixing proportions and/or densities. Bayesian Markov Chain Monte Carlo (MCMC) methods are often used to fit these models \citep{peng1996,wood2002,geweke2007}. FMMs require certain parameter specifications such as the mixing proportion values or number of densities which can affect their overall inference capabilities.

Infinite mixture models are another common Bayesian nonparametric mixture modeling approach. One class of infinite mixture model techniques attempts to directly estimate the conditional density via an infinite set of mixture weights and a process mixing distribution prior dependent on the covariates. \citet{dunson2007} develop a Bayesian density regression model using a local, covariate-weighted mixture of DP priors. \citet{trippa2011} and \citet{jara2011} propose use of a Polya Tree (PT) prior model and induce dependence through different definitions of the splitting probabilities. \citet{tokdar2010} forego these priors and develops a model using logistic Gaussian processes and subspace projection. Still, Bayesian non-parametric density estimation analysis can be computationally burdensome as data complexity increases, leading to some variable selection techniques being proposed \citep{chung2009,kundu2014}. Infinite mixture models for estimating the joint distribution of the response and covariates have also been proposed \citep{muller1996,shahbaba2009,park2010,taddy2010,hannah2011}. A disadvantage of this class of techniques is that it does not directly estimate the conditional density, and also can be slow in terms of computational performance as the dimensions of the problem increase.

Machine learning algorithms are another useful and arguably more accessible class of conditional density estimation methods. One approach is to use an orthogonal series density estimator that adapts to the geometric features of the data and reduces the dimension of the problem, with additional improvements later proposed via incorporation of regression and deep learning algorithms \citep{efromovich2010,izbicki2016,izbicki2017,dalmasso2020}. \citet{meinshausen2006} proposes a foundational method of quantile regression forest (QRF). By noting all observations in each leaf, a random forest can be used to calculate the full conditional distribution as a weighted sum of sample quantiles across trees. Multiple conditional density estimation methods using random forests to improve on QRF accuracy and/or speed have been developed \citep{tung2014,hothorn2017,pospisil2018}. Recently, \citet{li2019} proposed deep distribution regression (DDR) as a deep network learning-based conditional distribution technique. \citet{li2019} use cutpoints to discretize the response space and apply a multi-class classification method (such as a neural network) on the resulting bins. \citet{li2019} also give an approach which accounts for bin ordering by applying a binary classification model for each cutpoint and jointly estimating the conditional cumulative distribution function.

Similar to DDR, we consider a conditional density estimation approach to incorporate machine learning algorithms. A logistic transformation is made on the model output layer to obtain an expression of the conditional density function. The flexibility of the model specification allows for algorithms such as polynomial regression or deep learning models to be used. Our method evaluates only a single set of model parameters and simultaneously estimates the full conditional distribution. This information sharing allows our method to perform well when minimal data is available, and the relatively limited number of parameters needing to be estimated ensure computational speed for the polynomial regression model choice. The gradient calculation can quickly become intractable for complex model choices, so we incorporate theory from ecological and epidemiological statistics. \citet{fithian2013} review models that can be used to evaluate presence-only survey data, including the inhomogenous Poisson process (IPP) model. We adapt the IPP framework to our data setting to justify a discrete approximation of our method for computational purposes. We also justify a special case of this method through a matched case-control context to further increase computational efficiency \citep{jarner2002}.

After a review of the method and some potential model choices, we discuss the computational considerations for its implementation. Following this, the methodological strengths and weaknesses of our method are explored with a simulation and forecasting case study, with the takeaways and next steps summarized in a discussion section.

\section{Methods} \label{section_methods}

We are interested in approximating the conditional distribution of response variable $\Y\in\mathbb{R}$ given the covariate information $\bX\in\mathbb{R}^{p}$, denoted $\conddistfxnyX$. Our method requires a lower and upper bound for the target variable, which we address through a transformation of the response variable onto the unit interval. Suppose we transform $\Y$ through a cumulative distribution function $G$ as $\Z=G(\Y|\bX)\in[0,1]$. Note that the transformation of $\Y$ into $\Z$ to be on the unit interval is not unique, we could instead determine an upper and lower interval bound for $\Y$ on its original scale.

In this section, we will outline our method for approximating the conditional distribution of the transformed response, $\conddistfxnzX$, however the conditional density of the original $\conddistfxnyX$ can be recovered applying the change of variable formula as
\begin{align}
		\conddistfxnyX=f(G(\y)|\bX)\bigg| \frac{\partial G}{\partial \y} G(\y) \bigg|. \label{change_of_variable_formula}
\end{align}
If $\conddistfxnzX$ is uniformly distributed, the resulting $\conddistfxnyX$ distribution will be governed by $G$. In other words, $G$ is the base predicted distribution family, as opposed to the uniform distribution if no transformation of $\Y$ is made.

\subsection{Logistic Transformation}

Let $\qzX$ be a smooth function over $\z$ and $\bX$. The logistic transformation (e.g. \citet{lenk1988}) relates $\qzX$ to $\conddistfxnzX$ as

\begin{align}
	\conddistfxnzX=\frac{e^{\qzX} }{ \int_{0}^{1} e^{\quX}d\integralstdresponse }. \label{logistictransformationcontinuous}
\end{align}

\noindent Since $\qzX=A(\z,\bX)+B(\bX)$ gives the same density as $\qzX=A(\z,\bX)$, the main effect terms for $\bX$ are removed. As the support of $\qzX$ is arbitrarily flexible, any smooth conditional probability density function $\conddistfxnzX$ can be modeled with this transformation. In practice, this integral may be intractable. Discrete approximation techniques are discussed in \ref{computing_subsection} after introducing potential model choices.

A smooth underlying $\q$ function allows for the simultaneous estimation of a single set of model parameters. 
A similar logistic transformation on an underlying model was used in \citet{tokdar2012} to develop a simultaneous quantile regression estimation method. The information sharing inherent in this approach enabled estimation of multiple quantiles concurrently, improving on previous quantile regression estimation methods.

Another advantage of this method is its flexibility. The only required $\q$ function specification is smoothness, which allows for many non-parametric model possibilities. We consider two such models in this paper which draw from machine learning ideas, a polynomial regression model and a deep learning model. However, our method can easily be applied to other smooth model choices such as an additive model with splines.

\subsection{Polynomial Regression Model} \label{polynomialregressionmodel_subsection}

The Weierstrass Approximation Theorem states that for any continuous real-valued function on a closed interval, there exists a polynomial function that can approximate it arbitrarily well \citep{weierstrass1885}. The polynomial function is therefore a logical candidate for the smooth function in our method. Let $\polynomialpowers$ be an integer representing the largest polynomial power used for the centered $\Z$ values, with $\polynomialpower$ representing the given polynomial power. Recall $\cov=1,...,\covs$ represents the covariate. Also, let $\maxquadraticterm=1,...,\maxquadraticterms$ index the polynomial degree associated with the covariate terms. We let $\maxquadraticterm=2$ and give the second-order model as

\begin{align}
\qzX=\sum\limits_{\polynomialpower=1}^{\polynomialpowers}\bigg[(\z-.5)^\polynomialpower\polynomialmodelparam_{\polynomialpower0}+\sum\limits_{\cov=1}^{\covs}\sum\limits_{\maxquadraticterm=1}^{2}(\z-.5)^\polynomialpower\X_{\cov}^{\maxquadraticterm}\polynomialmodelparam_{\polynomialpower\cov\maxquadraticterm} + \sum\limits_{\cov\neq\covtwo}(\z-.5)^\polynomialpower\X_{\cov}\X_{\covtwo}\polynomialmodelparamtwo_{\polynomialpower l}\bigg] \label{polynomialregressionmodel}
\end{align}

\noindent where $\polynomialmodelparam_{\polynomialpower0}$ represents the intercept, $\polynomialmodelparam_{\polynomialpower\cov\maxquadraticterm}$ represent the covariate coefficients, and $\polynomialmodelparamtwo_{\polynomialpower l}$ represent the $l$ interaction term coefficients. A higher order model follows this structure in the obvious way. The terms are centered by subtracting $0.5$ to reduce collinearity, and the main effects of $\bX$ are removed because they do not affect the conditional distribution.

\subsection{Deep Learning Model} \label{neuralnetworkmodel_subsection}

A deep learning model is another natural choice for the underlying smooth function. The universal approximation theorem states that a feed-forward artificial neural network with at least one hidden layer can approximate a continuous function on a compact space arbitrarily well \citep{hornik1989}. We propose a deep learning model with an input layer, at least one hidden layer, and an output layer. One hidden layer is given here for notational simplicity, but additional layers could be added if desired. Let $\outputlayercoef, \hiddenlayercoef$, and $\inputlayercoef$ represent the output layer, hidden layer, and input layer parameters, respectively. Let $\hiddenlayernode$ and $\inputlayernode$ represent the output and hidden layer nodes, respectively. Lastly, let $\inputlayernodeindex=1,...,\inputlayernodeindexes$ and $\hiddenlayernodeindex=1,...,\hiddenlayernodeindexes$ index the number of neurons in the hidden and output layer, respectively. The model is

\begin{align}
			\qzX=&\sum\limits_{\hiddenlayernodeindex=1}^{\hiddenlayernodeindexes} \outputlayercoefi \activationfunction(\hiddenlayernode_{\hiddenlayernodeindex}), \\
			\hiddenlayernode_{\hiddenlayernodeindex}=&\hiddenlayercoef_{0\hiddenlayernodeindex}+\sum\limits_{\inputlayernodeindex=1}^{\inputlayernodeindexes} \hiddenlayercoefi \activationfunction(\inputlayernode_{\inputlayernodeindex}), \\
			\inputlayernode_{\inputlayernodeindex}=&\inputlayercoef_{0\inputlayernodeindex}+\inputlayercoef_{1\inputlayernodeindex}(\z-.5)+\sum\limits_{\cov=2}^{\covs+1}\inputlayercoefi\X_{\cov},
\end{align}

\noindent where $\activationfunction$ is an activation function. Exponential linear unit (ELU) or rectified linear unit (ReLU) are two possible activation function options.

\section{Computing} \label{computing_subsection}

\subsubsection{Inhomogenous Poisson Process (IPP) Approximation} \label{ipp_approximation_subsubsection}

The only restriction for $\qzX$ is that it is smooth, potentially allowing the model to be highly complex. This model specification flexibility is an appealing feature, but can make the integral in the logistic transformation intractable. We can view our method in an inhomogenous Poisson process (IPP) model framework to justify a discrete logistic transformation which is more computationally feasible. The conditional density in \ref{logistictransformationcontinuous} has the form of an IPP model with domain on the unit interval $[0,1]$ and log-intensity $\qzX$. 
 
\citet{fithian2013} describe a discrete approximation of the IPP model, which we can apply to our context. Suppose we have a dataset with $\obs=1,...,\obss$ observations. We let $\ProjectTwozi$ denote the transformed response value for observation $\obs$. We can view the univariate random variable $\Z$ conditioned on $\bX$ as a location on the unit interval, so we can consider the observed data as realizations of a point process over the unit interval. We follow the IPP approximation literature and propose to approximate the likelihood contribution of observation $\obs$ as

\begin{align}
	\conddistfxnziX\approx\frac{e^{\qziX} }{ e^{\qziX}+\sum\limits_{\control=1}^{\controls} e^{\q(\z^*_{\obs\control},\bX)} }\label{conditionaldistributionmodel_data}
\end{align}

\noindent for $\control=1,...,\controls$. $\zilstar\sim\text{Uniform}(0,1)$ controls are uniquely selected for each observation. \citet{fithian2013} argue that this Monte Carlo approximation to the denominator of \ref{logistictransformationcontinuous} is accurate for sufficiently large $\controls$ in terms of approximating continuous conditional densities. The main effects of $\bX$ are removed for this discrete logistic transformation just as they were in \ref{logistictransformationcontinuous}. $\zilstar$ can instead be selected using a fixed grid across the unit interval, but we expect this choice would require a larger $\controls$ unless the data is evenly spread across the response space. This even data spread is the motivation for our transformation of $\Y$ by a CDF function, as a well-defined CDF can render the transformed data roughly uniform across the unit interval.

Another view of \ref{conditionaldistributionmodel_data} is that $\ProjectTwozi$ represents a sample from the location distribution of cases and the $\zilstar$ represent $\controls$ matched samples from the uniform control distribution \citep{jarner2002}. As mentioned in Appendix \ref{appendix:a}, even a small $\controls$ provides valid information about the $\q$ function. Thus, we can consider either the IPP approximation with large $\controls$ to approximate the IPP integrated intensity and the matched case-control approximation where $\controls=1$. We expect that a larger $\controls$ value will induce more accurate parameter estimation, but at an additional computational cost that may not always be feasible.

Standard optimization methods can be employed with this approximation by minimizing the negative log likelihood objective function. Let $\paramvec\in\mathbb{R}^{\params}$ represent the parameter vector for the chosen $\q$ model, which we can write as $\qzXtheta$. The negative log likelihood for our model is

\begin{align}
	\ell(\paramvec|\ProjectTwoZi,\bXi)=\sum\limits_{\obs=1}^{\obss}\bigg\{-\qziXitheta+\log\bigg[e^{\qziXitheta}+\sum\limits_{\control=1}^{\controls} e^{\qzikstarXltheta} \bigg]\bigg\}+\ridgepenalty||\paramvec||^2 \label{negativeloglikelihoodmodel}
\end{align}

\noindent where $\ridgepenalty\geq0$ is a ridge penalty included to avoid model overfitting. For $\controls=1$, the method effectively reduces to logistic regression and the polynomial model can be evaluated using penalized logistic regression analysis techniques \citep{friedman2010}. This technique arrives at a solution extremely quickly, making the polynomial method very accessible for large datasets. For deep learning methods, we perform stochastic gradient descent. Details for these implementation choices can be found in Appendix \ref{appendix:b}.

Let $\testingresponsevalind=1,...,\testingresponsevalinds$ index a set of transformed response values. We can predict the conditional distribution at these transformed response values given covariate vector $\bX$ and estimated parameter vector $\paramvec=\estparamvec$ as

\begin{align}
	\conddistfxnzlXestparamvec\approx\frac{e^{\q(\z_\testingresponsevalind,\bX;\estparamvec)} }{ \sum\limits_{j=1}^{\testingresponsevalinds} e^{\qzjXthetahat} } . \label{testing_discreteconditionaldistribution}
\end{align}

This can be transformed back to the original scale via \ref{change_of_variable_formula}. A key advantage of our method is its simultaneous estimation of the model parameters. This structure ensures that we implicitly share information across all of our quantile estimates. For a method like DDR with a multinomial logistic regression classification model, each bin has its own set of parameters to be evaluated (excluding one bin which serves as a reference for the others). If a bin contains few or no observations, then that bin's parameter estimates may be volatile and unreliable. A large number of cut points may be desired to approximate a continuous distribution estimate, which makes it more likely there are empty or sparsely filled bins. Our method avoids this issue by estimating parameters for only a single model, implicitly assuring information is shared across all quantile estimates. For certain model choices, another benefit of this single set of model parameters is that our method becomes computationally quicker than DDR and even QRF.


\section{Simulation Study}

We conduct a simulation study to evaluate our method against the aforementioned DDR and QRF methods \citep{meinshausen2006, li2019}. We compare these three machine learning-based methods in terms of effectiveness in predicting the conditional distribution of the target variable, explained below.

We simulate data from four distributions, first used by \citet{li2019} for their complicated structures. Model 1 has a linear mean function, but also an error term that varies with the covariates. The other three models have a nonlinear mean function. Models 2 and 3 are mixture distributions, while Model 4 uses a skew-normal distribution for the errors. Formally, the models are specified as
 
\begin{itemize}
	\item Model 1: $ Y=\boldsymbol{X}^{T} \boldsymbol{\beta}_{1} +\exp \left(\boldsymbol{X}^{T} \boldsymbol{\beta}_{2}\right)* \epsilon$,
	\begin{itemize}
		\item $\boldsymbol{X}\sim\MVN(\bm{0},\bm{I_5})$,
		\item $\boldsymbol{\beta}_{1} \sim N\left(\mathbf{0}, \bm{I_{5}}\right)$, $\boldsymbol{\beta}_{2} \sim N\left(\mathbf{0}, 0.45 \bm{I_{5}}\right)$, $\epsilon \sim N(0,1)$.
	\end{itemize}
	\item Model 2: $Y=\left[10 \sin \left(2 \pi X_{1} X_{2}\right)+10 X_{4}+\epsilon_{1}\right] \pi_{1}+\left[20\left(X_{3}-0.5\right)^{2}+5 X_{5}+\epsilon_{2}\right]\left(1-\pi_{1}\right)$,
	\begin{itemize}
		\item $X_{1}, \cdots, X_{10} \stackrel{i i d}{\sim} \text {Uniform}(0,1)$,
		\item $\pi_{1} \sim \text { Bernoulli }(0.5)$, $\epsilon_{1} \sim N(0,2.25)$, $\epsilon_{2} \sim N(0,1)$.
	\end{itemize}

	\item Model 3: $Y=\left[\sin \left(X_{1}\right)+\epsilon_{1}\right] \pi_{1}+\left[2 \sin \left(1.5 X_{1}+1\right)+\epsilon_{2}\right]\left(1-\pi_{1}\right)$,
	\begin{itemize}
		\item $X_{1} \sim \text{Uniform}(0,10)$,
		\item $\pi_{1} \sim \text {Bernoulli}(0.5)$, $\epsilon_{1} \sim N(0,0.09)$, $\epsilon_{2} \sim N(0,0.64)$.
	\end{itemize}
	
	\item Model 4: $Y=10 \sin \left(2 \pi X_{1} X_{2}\right)+20\left(X_{3}-0.5\right)^{2}+10 X_{4}+5 X_{5}+\epsilon$,
	\begin{itemize}
		\item $X_{1}, \cdots, X_{10} \stackrel{i i d}{\sim} \text {Uniform}(0,1)$,
		\item $\epsilon \sim \text {SkewNormal}(0,1,-5)$.
	\end{itemize}
\end{itemize}

For each scenario, we simulate 100 datasets of size of 200, 1000, or 4000 observations to explore the relative efficacy of our method for various sample sizes. The datasets are randomly divided into training and testing data using a 75\%/25\% split. The models are fit using the training data, and then the distribution for each testing dataset observation is determined. For all models, the covariate data was normalized.

To evaluate the accuracy of a distribution estimate, we first calculate the range of the training response data and further extend it by 10\%. We then calculate 100 evenly-spaced cut points between the extended range boundaries. For each model, we calculate the empirical CDF value associated with every cut point to get the conditional distribution estimate for every observation. We use the divergence function associated with the continuous ranked probability score (CRPS) to evaluate method performance \citep{gneiting2007,kruger2016}. The CRPS divergence is defined as

\begin{align*}
d_{CRPS}=\frac{1}{N} \sum_{n=1}^{N} \int_{l}^{u}\left\{\hat{F}\left(y | \boldsymbol{X}_{n}\right)-F\left(y | \boldsymbol{X}_{n}\right)\right\}^{2} d y.
\end{align*}

\noindent This integral is approximated using $1000$ evenly gridded points and the resulting approximation is normalized by the range of the data. For the simulation study, $N$ denotes the number of testing set observations for the given scenario.

We apply a matched case-control (MCC) justified approximation with $\controls=1$ randomly selected controls to both the polynomial and deep learning models. Additionally, we apply an inhomogenous Poisson process (IPP) justified approximation with $\controls=10$ randomly selected controls to the deep learning model in the simulation scenarios with 200 observations. For the polynomial MCC approximation method, the first-order interaction terms between covariates and squared covariate terms were included in the covariate pool for Models 1, 2, and 4. For Model 3, there was only one covariate variable so no interaction terms were possible. The highest polynomial power used in the model was $\polynomialpowers=3$. 

Both deep learning approximations were applied using a model structure with one hidden layer. 30 nodes feed into the hidden and output layers each and the chosen activation function was the exponential linear unit (ELU). For the polynomial and deep learning methods, we select the normal cumulative distribution function (CDF) $\Phi$ to transform $\Y$ as $\Z=G(\Y|\bX)=\Phi\bigg(\frac{\Y-\bX\parquantbmeanparam}{\parquantsdparam}\bigg)$ and estimate mean coefficient $\parquantbmeanparam$ and standard deviation $\parquantsdparam$ parameters using ordinary least squares (OLS) regression. This choice ensures that the base distribution prediction for each observation in the testing dataset is Gaussian and centered at the OLS conditional mean. A larger ridge penalty (which lessens the deviance of the parameters from each other) will influence the predicted distribution toward this base distribution.

The polynomial MCC approximation is evaluated using a penalized logistic regression method while the deep learning approximations are evaluated using stochastic gradient descent. For more details on the implementation and evaluation of the models in these two methods, see Appendix \ref{appendix:b}.

The classification models for the DDR method were constructed using the deep-conditional-distribution-regression Python package found at https://github.com/RLstat/deep-conditional-distribution-regression. The joint binary cross entropy loss objective function was selected due to its superior performance over the multinomial objective function in \citet{li2019}. Models were built with a single hidden layer and a 0\% dropout rate. The ELU activation function is selected for the hidden layer, with a softmax activation function applied on the output layer.

The QRF method was utilized with 500 trees were built using the quantregForest package in R. This package predicts the conditional response values associated with inputted quantiles, so 100 evenly-spaced quantiles from .00001 to .99999 were generated and the QRF models estimated the cut points associated with these quantiles.

\begin{figure}[htbp!]
	\centering
		\includegraphics[width=1\textwidth]{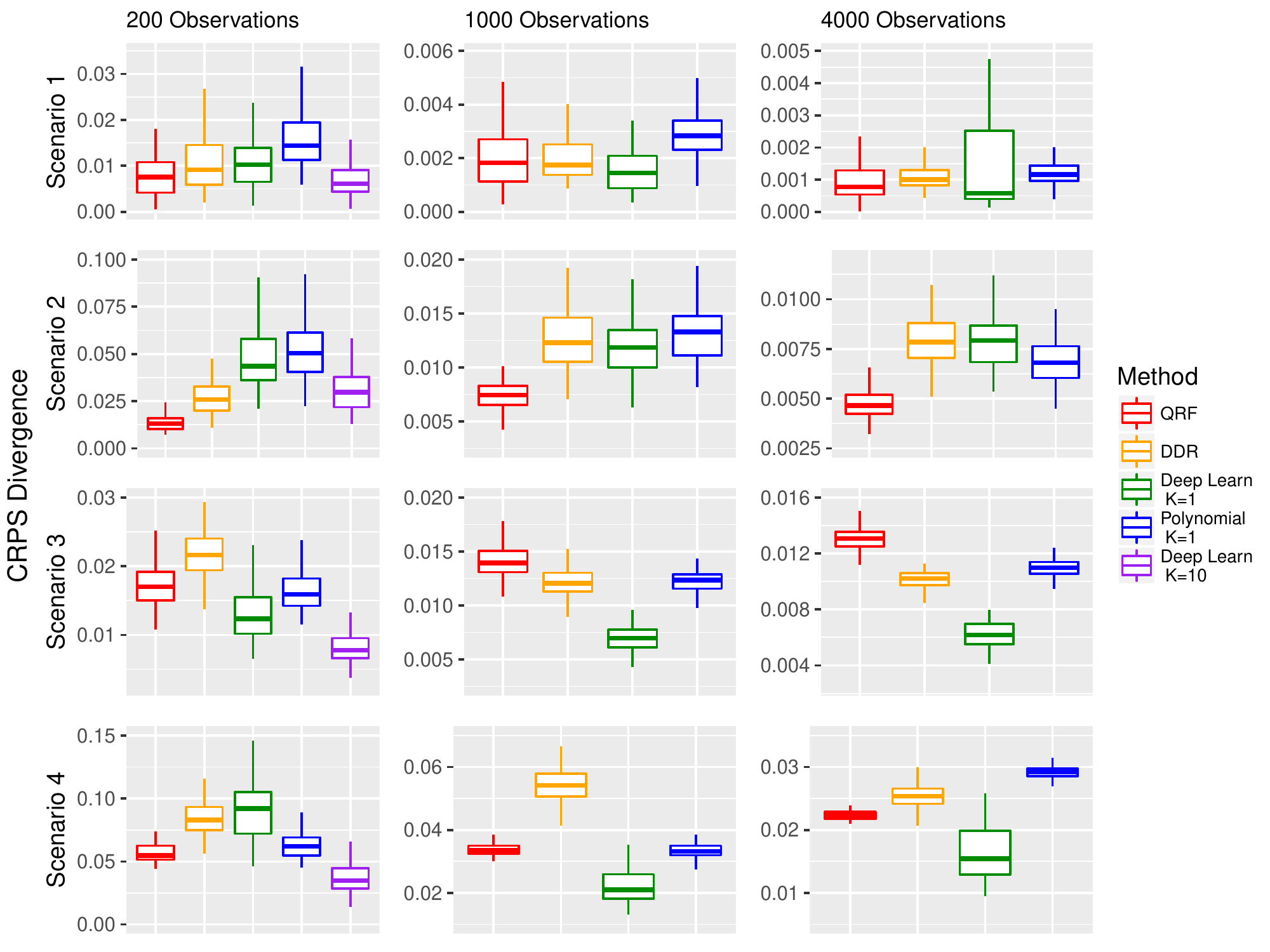}
		\caption{A boxplot of the distribution of CRPS divergences for each model and dataset size across 100 datasets for the QRF, DDR, Deep Learning $\controls=1$, Polynomial $\controls=1$, and Deep Learning $\controls=10$ conditional distribution estimation methods. The y-axis scale is not synchronized across scenarios and dataset sizes. \label{results_plot}}
\end{figure}

Figure \ref{results_plot} gives the simulation results. In general, both deep learning approximation methods performed well compared to DDR. The deep learning MCC approximation model outperformed DDR in terms of median CRPS divergence in 8 of the 12 data scenarios. The polynomial MCC approximation model performed worse against DDR by comparison, only producing a lower median CRPS divergence in 4 of the 12 scenarios. 

The deep learning IPP approximation method noticeably improved the CRPS divergence results compared to the deep learning MCC method in all four scenarios with 200 observations. In Models 1, 3, and 4 with 200 observations, this approximation beat both QRF and DDR in terms of median CRPS divergence, suggesting that this deep learning approximation is more useful than the MCC approximation in situations with a small sample size.

Our deep learning MCC approximation method also outperformed QRF in terms of median CRPS divergence in 7 of the 12 scenarios, although the relative CRPS divergence ranges in Model 1 with 4000 observations suggests our method may not have produced better results in that scenario. QRF fared better than the polynomial MCC approximation model in the majority of scenarios, although the polynomial MCC approximation produced lower median CRPS divergence values across dataset sizes in Model 3.

Our deep learning method performed relatively better in terms of CRPS divergence in Models 1 and 3 compared to Models 2 and 4. Model 1 had a normal distribution structure which may have been advantageous for our method since we used the normal quantile function to transform our data. Model 3 was a mixture distribution as Model 2 was, but only had a single covariate compared to the 10 covariates in Models 2 and 4.

Table \ref{computation_table} gives the average computation times for the polynomial and deep learning approximation methods for Model 1. The deep learning IPP approximation computation times for 1000 and 4000 observations were calculated on only 5 datasets, whereas the computation times for the other scenarios were calculated for all 100 datasets. The deep learning MCC and IPP approximations were significantly more computationally burdensome than the polynomial MCC approximation. The deep learning MCC approximation average computation time was over an hour for 4000 observations. On average, the deep learning IPP approximation for 10 controls took roughly five or six times as long to evaluate as the deep learning MCC approximation. Figure \ref{results_plot} suggests the deep learning MCC approximation and especially deep learning IPP approximation are preferable to the polynomial MCC approximation for conditional distribution estimation in many data scenarios, however it may not be as readily scalable to larger datasets. On the contrary, the increase in computation time from 200 observations to 4000 observations for the polynomial MCC approximation was negligible. The polynomial MCC approximation is easily applicable to large datasets in data scenarios where the deep learning approximations are computationally unfeasible.

\begin{table}[tp!]
\begin{tabular}{|l|c|c|c|}
\hline
\multicolumn{1}{|c|}{\multirow{4}{*}{Dataset Size}} & \multicolumn{3}{c|}{Mean Computation Time}     \\
\multicolumn{1}{|c|}{}                              & \multicolumn{3}{c|}{(Minutes) For Model 1}     \\ \cline{2-4} 
\multicolumn{1}{|c|}{}                              & Polynomial    & Deep Learning & Deep Learning  \\
\multicolumn{1}{|c|}{}                              & $\controls=1$ & $\controls=1$ & $\controls=10$ \\ \hline
\multirow{2}{*}{200}                                & 0.0002        & 3.067         & 16.740         \\
                                                    & (0.0002)      & (0.144)       & (0.213)        \\ \hline
\multirow{2}{*}{1000}                               & 0.0002        & 15.785        & 90.860         \\
                                                    & (0.0001)      & (0.303)       & (2.175)        \\ \hline
\multirow{2}{*}{4000}                               & 0.0005        & 67.531        & 373.713        \\
                                                    & (0.0001)      & (0.572)       & (5.358)        \\ \hline
\end{tabular}
\caption{A table of the average computation times (in minutes) and associated standard errors for evaluating the data from Model 1 across all dataset sizes. The average computation times were recorded for the polynomial MCC approximation ($\controls=1$), deep learning MCC approximation ($\controls=1$), and deep learning IPP approximation ($\controls=10$) conditional distribution estimation methods. The average computation times and standard errors for the deep learning IPP approximation method for 1000 and 4000 observations were calculated using only the first 5 datasets due to the burdensome evaluation time. The average computation times and standard errors for the remaining scenarios were calculated over all 100 datasets. The polynomial MCC approximation method was evaluated using penalized logistic regression, whereas the deep learning IPP approximation methods were evaluated using stochastic gradient descent. \label{computation_table}}
\end{table}


\section{Application To Tropical Cyclone Intensity Forecasting}\label{case_study}

We apply our method to calibrate short-term tropical cyclone wind intensity forecasts. A conditional distribution estimation approach to this problem could provide additional context on response distribution features to better inform policy decisions compared to a point estimate approach \citep{cloud2019}. Our data comes from Hurricane Weather Research and Forecasting (HWRF) Model, developed and maintained by the U.S. Environmental Modeling Center (EMC) \citep{biswas2017}. HWRF is a deterministic atmosphere-ocean model used for hurricane research and forecasting. The HWRF model includes a forecasted maximum 10-meter wind speed value which is designated as the covariate of interest. The actual maximum 10-meter wind speed value is the response variable. Covariate and response information are recorded up to four times a day for each day a tropical cyclone is active in 6 hour increments. At each time point, forecasted covariate data and response data are given for up to 96 hours into the future by 3 hour increments.

The full dataset contains information from 65 tropical cyclones located around the Atlantic Seaboard between 2013 and 2017. For this application, we focus on lag 3 and lag 6 forecast predictions and subset the overall dataset of 45,639 observations to obtain two smaller datasets of 1,383 observations each for only these lag times. Observations with missing response values were removed. The final lag 3 and lag 6 datasets each had 1,267 observations.

The polynomial regression method was implemented using the MCC approximation with a single control, $\controls=1$. The highest polynomial power used in the polynomial model was $\polynomialpowers=3$, and the quadratic covariate term was included in the covariate matrix. The deep learning method was implemented using an IPP approximation with $\controls=20$. The deep learning model was built with a single hidden layer, where 15 nodes feed into the hidden and output layers each, and uses an ELU activation function. The polynomial model was evaluated using penalized logistic regression and the deep learning model was evaluated using mini-batch stochastic gradient descent. For both methods, a variety of ridge penalties were considered. A ridge penalty of $0.000001$ was selected for the deep learning method for both lags and the polynomial method for lag 3, and a ridge penalty of $0.0005$ was selected for the polynomial method for lag 6. For the deep learning method, a variety of initial learning rates were also considered, with the optimally tuned models using an initial learning rate of $1$ for both lags. Further details on how these models were fit are given in Appendix \ref{appendix:b}. 

The QRF model was built using 500 trees and evaluated using the quantregForest R package. The DDR method was run using a deep learning classification model and evaluated in Python using the deep-conditional-distribution-regression package. The model had one hidden layer with 15 nodes and a 0\% dropout rate to mimic the deep learning approximation model specifications. The joint binary cross entropy loss objective function was selected. The ELU activation function was applied to the hidden layer, with the softmax activation function used for the output layer. As in the simulation study section, we select a normal CDF to transform $\Z$ and estimate the CDF parameters using OLS regression. 

The tropical cyclones were randomly assigned to one of five folds, and 5-fold cross validation was performed. For each fold, we calculate the CRPS of the testing set to evaluate method performance as the CRPS divergence is unavailable without knowledge of the true distribution \citep{matheson1976,hersbach2000}. CRPS is defined as

\begin{align*}
CRPS=\frac{1}{N} \sum_{n=1}^{N} \int_{l}^{u}\left\{\hat{F}\left(y | \boldsymbol{X}_{n}\right)-I\left(y \geq Y_{n}\right)\right\}^{2} d y. \label{CRPS_formula}
\end{align*}

\noindent As with the CRPS divergence evaluation, the integral is approximated using $1000$ evenly-gridded points and the resulting approximation is normalized by the range of the data. For this application, $N$ refers to the number of observations in the given testing fold.

Table \ref{case_study_results_table} gives the average CRPS across folds and the accompanying standard error for each method. Our polynomial and deep learning approximation methods outperform QRF and DDR by these metrics. Additionally, the deep learning IPP approximation slightly outperforms the polynomial MCC approximation in terms of average CRPS. The lag 6 predictions result in a higher average CRPS for each method than the lag 3 predictions, due to the increased difficulty of forecasting further into the future.

\begin{table}[tp!]
\begin{tabular}{|l|l|l|l|l|}
\hline
\multicolumn{1}{|c|}{\multirow{2}{*}{\textbf{Response Lag Time}}} & \multicolumn{4}{c|}{\textbf{Method}}              \\ \cline{2-5} 
\multicolumn{1}{|c|}{}                                            & QRF      & DDR      & Polynomial & Deep Learning \\ \hline
3 Hour Lag                                                             & 0.0247   & 0.0288    & 0.0218      & 0.0212          \\
                                                                  & (0.0037) & (0.0096) & (0.0023)   & (0.0022)       \\ \hline
6 Hour Lag                                                             & 0.0293   & 0.0299    & 0.0267      & 0.0258          \\
                                                                  & (0.0043) & (0.0070) & (0.0032)   & (0.0033)       \\ \hline
\end{tabular}
\caption{The 5-fold mean CRPS values for each lag time for the QRF, DDR, polynomial MCC approximation ($\controls=1$), and deep learning IPP approximation ($\controls=20$) conditional distribution estimation methods. \label{case_study_results_table}}
\end{table}

\begin{figure}[htbp!]
	\centering
		\includegraphics[width=1\textwidth]{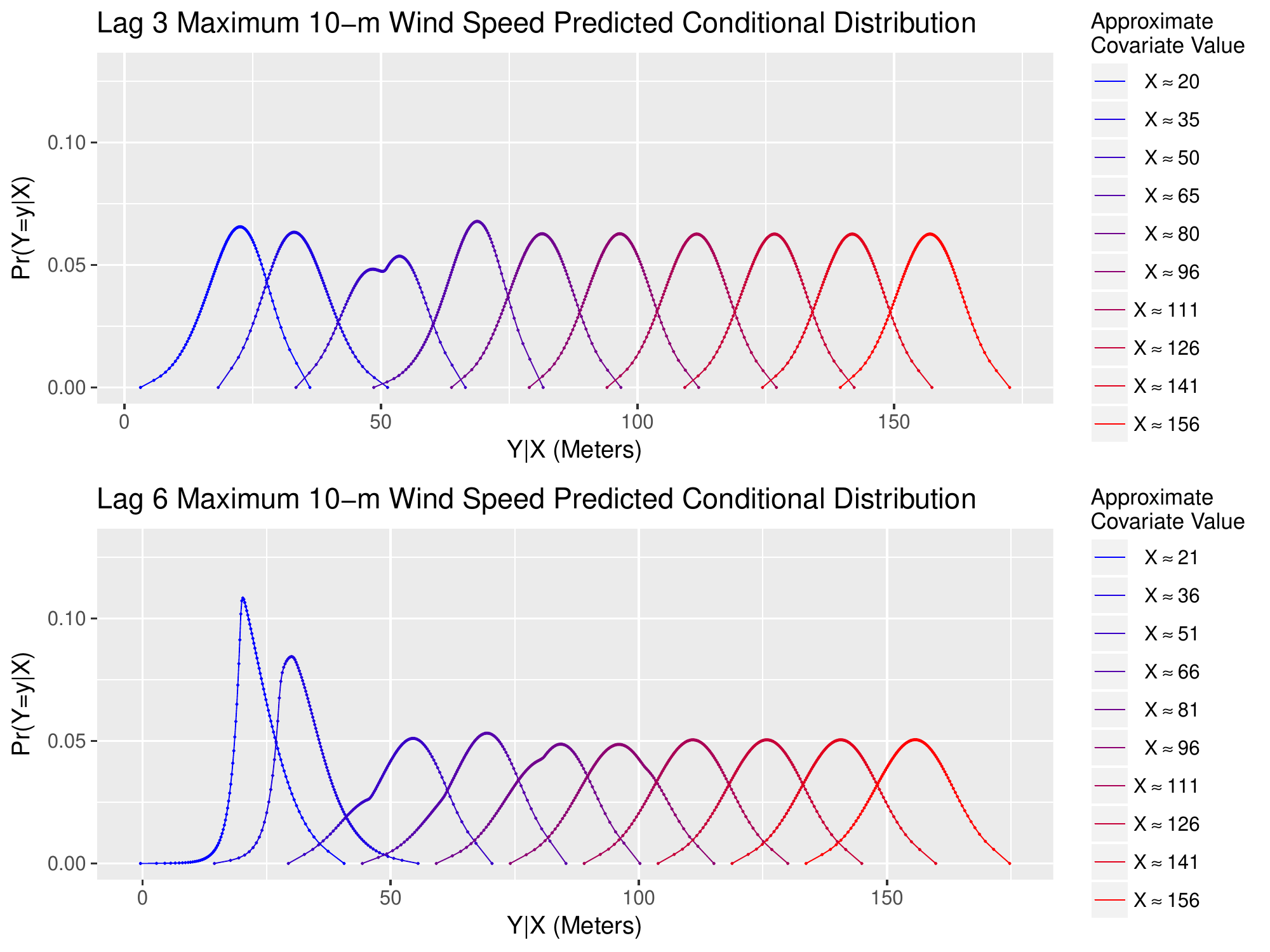}
		\caption{Deep learning IPP approximation ($\controls=20$) conditional maximum 10-meter wind speed distribution predictions for lag 3 and 6 using model constructed with all available data. Estimated conditional response probabilities for 100 equally spaced quantiles sequenced between $0.5$ and $99.5$ are displayed with linear interpolation between quantiles. The $0.5$th and $99.5$th quantile density function values are rounded to 0. $\Pr(\Y=\y|\X)$ refers to the relative probability that the maximum 10-meter wind speed $\Y=\y$ occurs given the HWRF-forecasted maximum 10-meter wind speed value $\X$. $\Y|\X$ refers to the conditional response value $\Y$ given the covariate $\X$. \label{CDE_application_plot}}
\end{figure}

\begin{figure}[htbp!]
	\centering
		\includegraphics[width=1\textwidth]{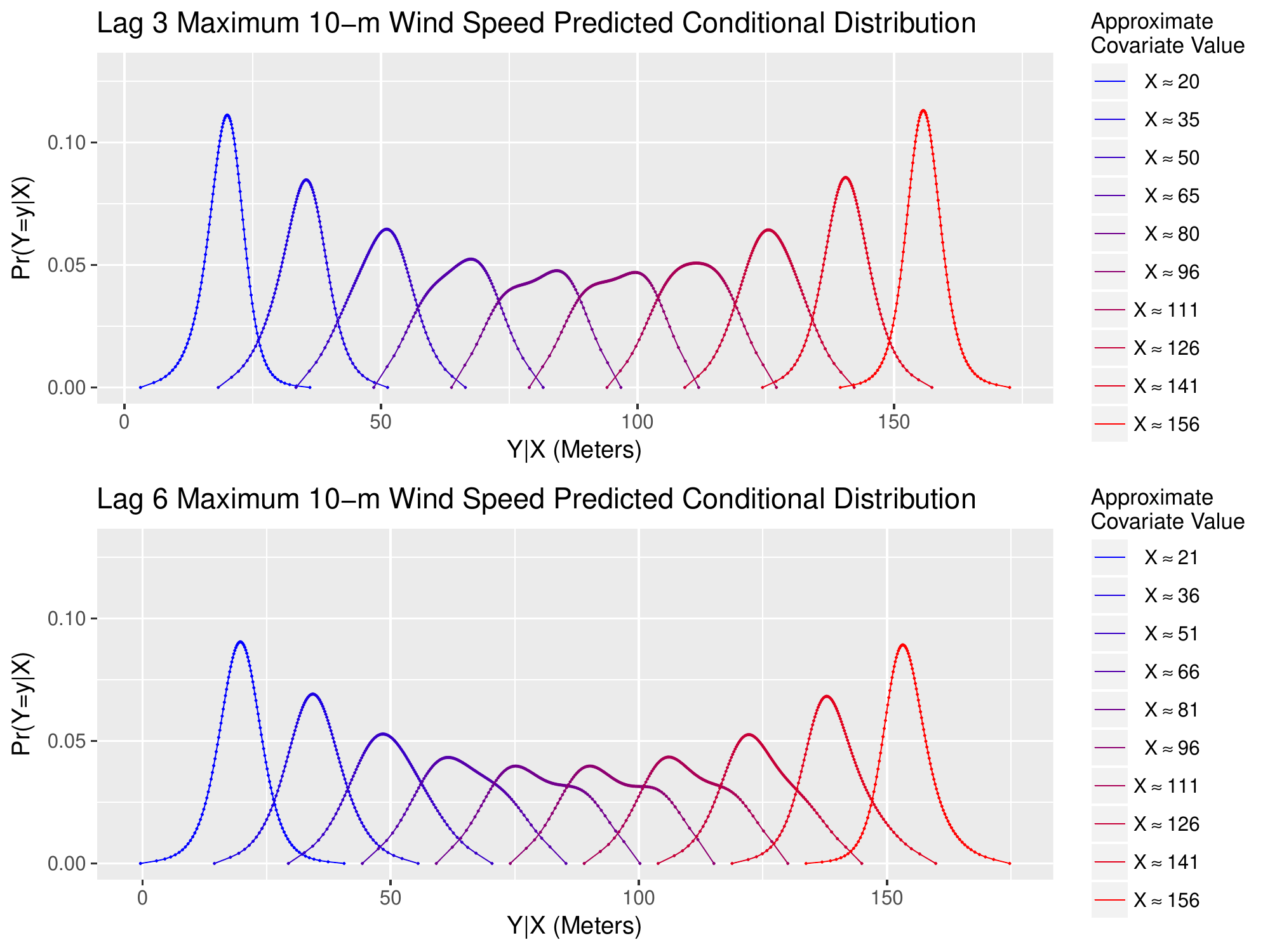}
		\caption{Polynomial MCC approximation ($\controls=1$) conditional maximum 10-meter wind speed distribution predictions for lag 3 and 6 using model constructed with all available data. Estimated conditional response probabilities for 100 equally spaced quantiles sequenced between $0.5$ and $99.5$ are displayed with linear interpolation between quantiles. The $0.5$th and $99.5$th quantile density function values are rounded to 0. $\Pr(\Y=\y|\X)$ refers to the relative probability that the maximum 10-meter wind speed $\Y=\y$ occurs given the HWRF-forecasted maximum 10-meter wind speed value $\X$. $\Y|\X$ refers to the conditional response value $\Y$ given the covariate $\X$. \label{Poly_CDE_application_plot}}
\end{figure}

As a comparison, the 5-fold mean CRPS value for the conditional Gaussian distribution evaluated via OLS estimation was calculated. The OLS-predicted mean CRPS values were $0.0220$ (with standard error $0.0020$) and $0.0269$ (with standard error $0.0026$) for lag 3 and lag 6, respectively. The deep learning IPP approximation and polynomial MCC approximation both outperform these estimated average CRPS values, although the improvement made by the polynomial MCC approximation is very slight.

Figure \ref{CDE_application_plot} displays the deep learning IPP approximation conditional response distribution predictions for lag 3 and lag 6 when using all of the training data. The lag 3 predicted quantiles for this model look generally unimodal and Gaussian, with some slight left skewness for smaller covariate values. The lag 6 predicted quantiles are also generally unimodal and Gaussian for the larger covariate values, but exhibit some clear non-normality and skewness for the lower covariate values. Both models used for Figure \ref{CDE_application_plot} were fit using the same ridge penalties and initial learning rates as the models used to calculate the CRPS values for each fold in Table \ref{case_study_results_table}. The predicted distributions for each individual fold that were used to calculate the mean CRPS results in Table \ref{case_study_results_table} are not necessarily equivalently shaped to these plotted predicted distributions. For instance, the deep learning IPP approximation method for lag 6 in the third fold predicts a somewhat bimodal distribution of maximum 10-m wind speed for larger covariate values. For an example plot of the predicted distributions using both methods and lag times for an individual fold, see Appendix \ref{appendix:c}. Overall, these somewhat Gaussian-shaped predicted distribution visualizations are consistent with the CRPS results that suggest the OLS-predicted distribution method performs only slightly worse than the deep learning IPP approximation for $\controls=20$ in this application.

Figure \ref{Poly_CDE_application_plot} displays the polynomial MCC approximation conditional response distribution for lag 3 and lag 6 when using all of the training data. Again, the models used to predict these distribution quantiles maintained the same ridge penalties and initial learning rates as the corresponding models used to obtain the polynomial MCC approximation average CRPS in Table \ref{case_study_results_table}. The larger and smaller covariate values are associated with distributions with sharper peaks, whereas the predicted distributions for the middle covariate values have broader, less symmetric peaks. The lag 3 and lag 6 predicted distributions look more similar here than the lag 3 and lag 6 predicted distributions in Figure \ref{CDE_application_plot}.

\section{Discussion}

We propose a flexible conditional distribution estimation method which can incorporate machine learning techniques such as deep learning models or polynomial regressions. We examined the performance of some of these model types for different data distributions in a simulation study, finding that our method implemented with a deep learning model outperformed other conditional distribution estimation methods in multiple scenarios. In a real world application of our method, we found both the deep learning and polynomial model-based methods provided useful insight on tropical cyclone maximum wind speed forecasting compared to other methods, with the deep learning method performing best in terms of the mean 5-fold CRPS performance metric.

Further approximation and/or computational techniques for this method can fully unlock its utility for conditional distribution estimation. We introduced an IPP-based discrete approximation with $\controls$ controls to make model evaluation feasible, but were limited to selecting a small $\controls$ and a relatively basic deep learning model structure with one hidden layer and 30 nodes. We expect that our method could substantially improve its predictive accuracy if a more complex deep learning model structure were tractable. Integration of our method with TensorFlow or another deep learning optimization programming language could be helpful in this regard. Perhaps an approximation that reduces the number of observations could also be incorporated to improve methodological accuracy.

Another potential methodological improvement is through the selection of the control values for our case-control based approximation. \citet{fithian2014} describe a local case-control sampling technique meant to address conditional imbalance in addition to the marginal imbalance issue addressed by standard case-control sampling. Perhaps incorporating this approach or another weighted control selection technique could be adapted to our framework to improve the conditional distribution estimation for a smaller number of controls.

For complicated models with many parameters (from multiple covariates, layers, and/or nodes), the ridge penalty in \ref{negativeloglikelihoodmodel} influences the parameters towards 0 so that they deviate less from each other. As a result, $g(\Z|\bX)$ tends toward a uniform distribution, and $f(\Y|\bX)$ consequently tends toward the distribution implied by the transformative cumulative distribution function. In both the simulation and application, a Gaussian cumulative distribution function was selected to transform $\Y$ to $\Z$. A conditionally normal response distribution is often assumed in statistics, so this specification is reasonable for many applications. Still, perhaps a more sophisticated optimization algorithm would allow for more deviance between parameter estimates and be less influenced by the choice of CDF.

Additionally, the CDF function parameters were estimated via OLS. OLS requires there to be more observations than covariates in order to obtain a unique parameter solution. This restriction might disallow the inclusion of higher order interactions in the polynomial approximation model when the sample size is small because it would result in more covariates than training observations. In this scenario, the CDF function transformation should not be used. Instead, boundaries should be chosen for $\Y$ and our method can be analogously applied.

\section{Acknowledgements}

We'd like to thank Dr. Kevin Gunn for lending equipment and technical assistance to obtain some of the results for the tropical cyclone intensity forecasting application.




\begin{appendices}


\section{Case-Control Sampling Justification} \label{appendix:a}

We exploit the relationship between an inhomogenous Poisson process (IPP) model and our problem to re-frame case and control samples into presence and background samples from presence-only datasets. Presence-only data is generally used with species distribution modeling where surveyors record all species presences within a pre-specified region along with randomly sampled background data across the region. \citet{fithian2013} lay out how to model presence-only data in a two-dimensional spatial domain using an IPP model. For our method, the domain for cases (and controls) is $[0,1]$ as previously noted. The IPP requires an intensity function $\intensityfxn$ to be specified which represents the likelihood of a case being present at any location in the given domain. The average intensity function for response value $\z$ over the domain is given as

\begin{align}
\IntensityFxn=\int_{0}^{1}\intensityfxn(\z)\dz.
\end{align}

An IPP model with intensity function $\intensityfxn$ gives the probability distributions for both the total number of cases as well as the locations of those cases. Conditional on the number of cases (governed by a Poisson distribution with mean $\IntensityFxn$), the locations of the cases are independently and identically distributed as

\begin{align}
\Pr(\Z=\z)=\frac{\intensityfxn(\z)}{\IntensityFxn}.
\end{align}

If we define $\intensityfxn(\z)=e^{\q(\z,\bX)}$, then the IPP model becomes

\begin{align}
\Pr(\Z=\z)=\frac{e^{\q(\z,\bX)}}{\int_0^1 e^{\q(\z,\bX)} \dz } .
\end{align}

We recognize this distribution form in the continuous logistic transformation formula given in \ref{logistictransformationcontinuous}. The cases are independently and identically distributed, so the log likelihood objective function is

\begin{align}
\ell(\paramvec|\Z,\bX)=\sum\limits_{\obs=1}^{\obss} \bigg[ \qziXitheta-\log\bigg( \int_0^1 \qzXitheta \dz \bigg) \bigg]
\end{align}

\noindent where $\obs=1,...,\obss$ and $\paramvec$ is the parameter vector for the selected $\q$ model, as before. \citet{fithian2013} discuss how to evaluate the integral in the denominator by approximating it using a finite set of control (background) samples. Let $\control=1,...,\controls$ index the control samples, so that $\zik$ denotes the $\control$th control value for observation $\obs$. Then, the log likelihood objective function with an approximated integral becomes
 
 \begin{align}
\ell(\paramvec|\Z,\bX)=\sum\limits_{\obs=1}^{\obss}\bigg[ \qziXitheta-\log\bigg( e^{\qziXitheta}+ \sum_{\control=1}^{\controls} e^{\qzikXitheta} \bigg) \bigg].
\end{align}

We recognize this as the log likelihood objective function in \ref{negativeloglikelihoodmodel}. Fithian and Hastie do not approximate the integral using a $\q$ function with a case plus a set of controls, instead only using the set of controls. However the case is only providing more information about the integral than the controls would on their own so we do not expect this to produce any inconsistencies or biases. They state that the control points can be uniformly sampled from the domain, which implies that selecting as few as $\controls=1$ controls still can provide valid inference about the target distribution function. Fithian and Hastie also state that the control points can be chosen through weighted sampling, referring to using quadrature weights. The additional ridge penalty component in \ref{negativeloglikelihoodmodel} could have been added to the IPP likelihood as well.

Note that for presence-only data, Fithian and Hastie detail possible sampling bias issues that can arise due to imperfect detection of presences during data collection. However, we do not think detectability issues are relevant to our usage of this framework. We make the assumption that the detectability parameter equals 1 in our context, so we do not have to worry about sampling bias in this respect.


\section{Computation} \label{appendix:b}

We use a variety of techniques to determine the optimal set of parameter estimates for the chosen model.

\subsection{Polynomial Regression Model} \label{appendix_computation_polynomial}

For the polynomial regression model, we can manipulate the data in order to evaluate the parameters using penalized logistic regression. Recall the polynomial regression model in \ref{polynomialregressionmodel}. We can express the polynomial model in \ref{polynomialregressionmodel} in matrix form as $\qzX=\bX(\z,\polynomialpowers)\polynomialmodelparamvec$, where $\bX(\z,\polynomialpowers)$ is the full covariate matrix as a function of the transformed response $\z$ and the highest polynomial power $\polynomialpowers$ and $\polynomialmodelparamvec=[\polynomialmodelparamvec_{10},,...,\polynomialmodelparamvec_{{\polynomialpowers}\covs} ]$ is the associated parameter vector. Then, we can rewrite the objective function in \ref{negativeloglikelihoodmodel} using the polynomial regression from \ref{polynomialregressionmodel} as

\begin{align}
\ell(\paramvec|\ProjectTwoZi,\bXi)=&\sum\limits_{\obs=1}^{\obss}\bigg\{ -\log\frac{e^{\bXi(\ProjectTwozi,\polynomialpowers)\polynomialmodelparamvec}}{e^{\bXi(\ProjectTwozi,\polynomialpowers)\polynomialmodelparamvec}+e^{\bXi(\zionestar,\polynomialpowers)\polynomialmodelparamvec}} \bigg\}+\ridgepenalty||\paramvec||^2 \\
=&\sum\limits_{\obs=1}^{\obss}\bigg\{ -\log\frac{1}{1+e^{[\bXi(\zionestar,\polynomialpowers)-\bXi(\ProjectTwozi,\polynomialpowers)]\polynomialmodelparamvec}} \bigg\}+\ridgepenalty||\paramvec||^2. \label{casecontrolapproximatednegativeloglikelihood_rewritten}
\end{align}

We recognize the form of this objective function for a logistic regression where every binary outcome equals 1. If we add a small amount of dummy observations where the covariate and outcome values are all 0, we can evaluate our parameter vector using penalized logistic regression. For each run of the polynomial regression case-control method, we used two dummy observations. The GLMNet package in R evaluates these parameters extremely fast, making this method practical and convenient \citep{friedman2009}.

We ran preliminary simulation runs on a smaller number of datasets for a variety of ridge penalties to determine the appropriate penalty values to use for each scenario of the simulation and the application. For Scenario 1, 2, and 4 the ridge penalties used were $0.025$, $0.05$, and $0.05$, respectively. For Scenario 3, the ridge penalty used for 200 observations was $0.025$, whereas the ridge penalty used for the other two sample size settings was $0.01$. 

For the application, we considered a variety of ridge penalties to determine which penalty minimized the mean 5-fold CRPS. For lag 3 and lag 6, a ridge penalty of $0.000001$ and $0.0005$ were chosen, respectively.


\subsection{Deep Learning Model}

For the deep learning case-control approximation, we cannot manipulate the data like in \ref{casecontrolapproximatednegativeloglikelihood_rewritten} and instead rely on gradient descent to evaluate the parameters.

Deep learning models can be difficult to train with a basic gradient descent algorithm. We extend or alter the basic gradient descent algorithm in a multitude of ways designed to improve parameter estimation. Mini-batch gradient descent is used with batch sizes of 50 and an initial step size of 1. For the case study, the batch sizes are approximate as the number of observations is not divisible by 50. The models are run for 600 gradient descent steps for the simulation and 300 gradient descent steps for the case study, respectively. Adaptive moment estimation (ADAM) is implemented along with a step decay that halves the initial step size every 100 steps for the simulation models and 50 steps for the case study models \citep{kingma2014}. We apply the batch normalization algorithm described in \citet{ioffe2015}. Model weights are initiated using the He initialization scheme, while the batch normalization shift and scale parameters are initialized to 0 and 1, respectively \citep{he2015}.

The gradient descent algorithm requires the calculation of a gradient vector which contains the first derivative value of the objective function with respect to each parameter. We obtain these values using back-propagation. The individual chain rule components for the gradient vector calculations for a single observation and case/control value are given in \ref{backprop}. The gradient vector can be calculated by summing the appropriate terms across observations and cases/controls.

\begin{align}\label{backprop}
\frac{\partial \ell}{\partial \qziXitheta}=&\frac{1}{e^{\qziXitheta}+e^{\qzikstarXltheta}} \\ 
\frac{\partial \ell}{\partial \qzikstarXltheta}=&\frac{e^{\qziXitheta}}{e^{\qziXitheta}+e^{\qzikstarXltheta}}\nonumber \\ 
\frac{\partial \qziXitheta}{\partial \outputlayercoefi}=&\frac{\partial \qzikstarXltheta}{\partial \outputlayercoefi}=\activationfunction(\hiddenlayernode_{\hiddenlayernodeindex}) \nonumber\\
\frac{\partial \qziXitheta}{\partial \activationfunction(\hiddenlayernode_{\hiddenlayernodeindex})}=&\frac{\partial \qzikstarXltheta}{\partial \activationfunction(\hiddenlayernode_{\hiddenlayernodeindex})}=\outputlayercoefi \nonumber\\
\frac{\partial \activationfunction(\hiddenlayernode_{\hiddenlayernodeindex})}{\partial \hiddenlayernode_{\hiddenlayernodeindex}}=&
\begin{cases}
1 & \text{if } \hiddenlayernode_{\hiddenlayernodeindex}>0\\ 
\alpha\exp{\hiddenlayernode_{\hiddenlayernodeindex}} & \text{if } \hiddenlayernode_{\hiddenlayernodeindex}\leq0
\end{cases} \nonumber\\
\frac{\partial \hiddenlayernode_{\hiddenlayernodeindex}}{\partial \activationfunction(\inputlayernode_{\inputlayernodeindex})}=&\hiddenlayercoefi \nonumber
\end{align}
\begin{align*}
\frac{\partial \hiddenlayernode_{\hiddenlayernodeindex}}{\partial \hiddenlayercoef_{0\hiddenlayernodeindex}}=&1 \nonumber\\
\frac{\partial \hiddenlayernode_{\hiddenlayernodeindex}}{\partial \hiddenlayercoefi }=&\activationfunction(\inputlayernode_{\inputlayernodeindex}) \nonumber \\
\frac{\partial \activationfunction(\inputlayernode_{\inputlayernodeindex})}{\partial \inputlayernode_{\inputlayernodeindex}}
=&\begin{cases}
1 & \text{if } \inputlayernode_{\inputlayernodeindex}>0\\ 
\alpha\exp{\inputlayernode_{\inputlayernodeindex}} & \text{if } \inputlayernode_{\inputlayernodeindex}\leq0
\end{cases}\nonumber\\
\frac{\partial \inputlayernode_{\inputlayernodeindex}}{\inputlayercoef_{0\inputlayernodeindex}}=&1 \nonumber\\
\frac{\partial \inputlayernode_{\inputlayernodeindex}}{\inputlayercoef_{1\inputlayernodeindex}}=&\ProjectTwozi \nonumber\\
\frac{\partial \inputlayernode_{\inputlayernodeindex}}{\inputlayercoefi}=&\X_{\cov\obs}. \nonumber
\end{align*}

The components for the batch normalization parameters can be obtained following the steps in \citep{ioffe2015}.

We ran preliminary simulation runs on a smaller number of datasets for a variety of ridge penalties to determine the appropriate penalty values to use for each scenario of the simulation and the application. For simulation Scenario 1, we used $0.025$, $0.002$, and $0.0025$ as the penalties for 200, 1000, and 4000 observations respectively. For simulation Scenario 2, we used $0.015$, $0.0025$, and $0.001$ as the penalties for 200, 1000, and 4000 observations respectively. For simulation Scenario 3, we used $0.0075$, $0.001$, and $0.001$ as the penalties for 200, 1000, and 4000 observations respectively. For simulation Scenario 4, we used $0.01$, $0.001$, and $0.0005$ as the penalties for 200, 1000, and 4000 observations respectively.

For the application, we considered a variety of ridge penalties and initial learning rates for the deep learning IPP approximation method. For both lags, the tuning settings which optimized the mean 5-fold CRPS were a ridge penalty of $0.000001$ and an initial learning rate of $1$.



\section{Predicted Distribution Plots For Individual Fold} \label{appendix:c}

The predicted distribution plots for a range of covariate values using the full tropical cyclone dataset is given in Section \ref{case_study}, however the results in Table \ref{case_study_results_table} were produced by obtaining separate models for each testing dataset fold. Here, we present an example of the predicted distribution plots for the third fold for both methods and both lags. The predicted distributions for the individual folds can vary somewhat compared to the overall predicted response distributions fitted using the full tropical cyclone dataset, as seen below for the deep learning IPP approximation. The polynomial MCC approximation predicted conditional distributions for the individual folds were generally similar to the overall predicted conditional distributions in Figure \ref{Poly_CDE_application_plot}.

\begin{figure}[htbp!]
	\centering
		\includegraphics[width=1\textwidth]{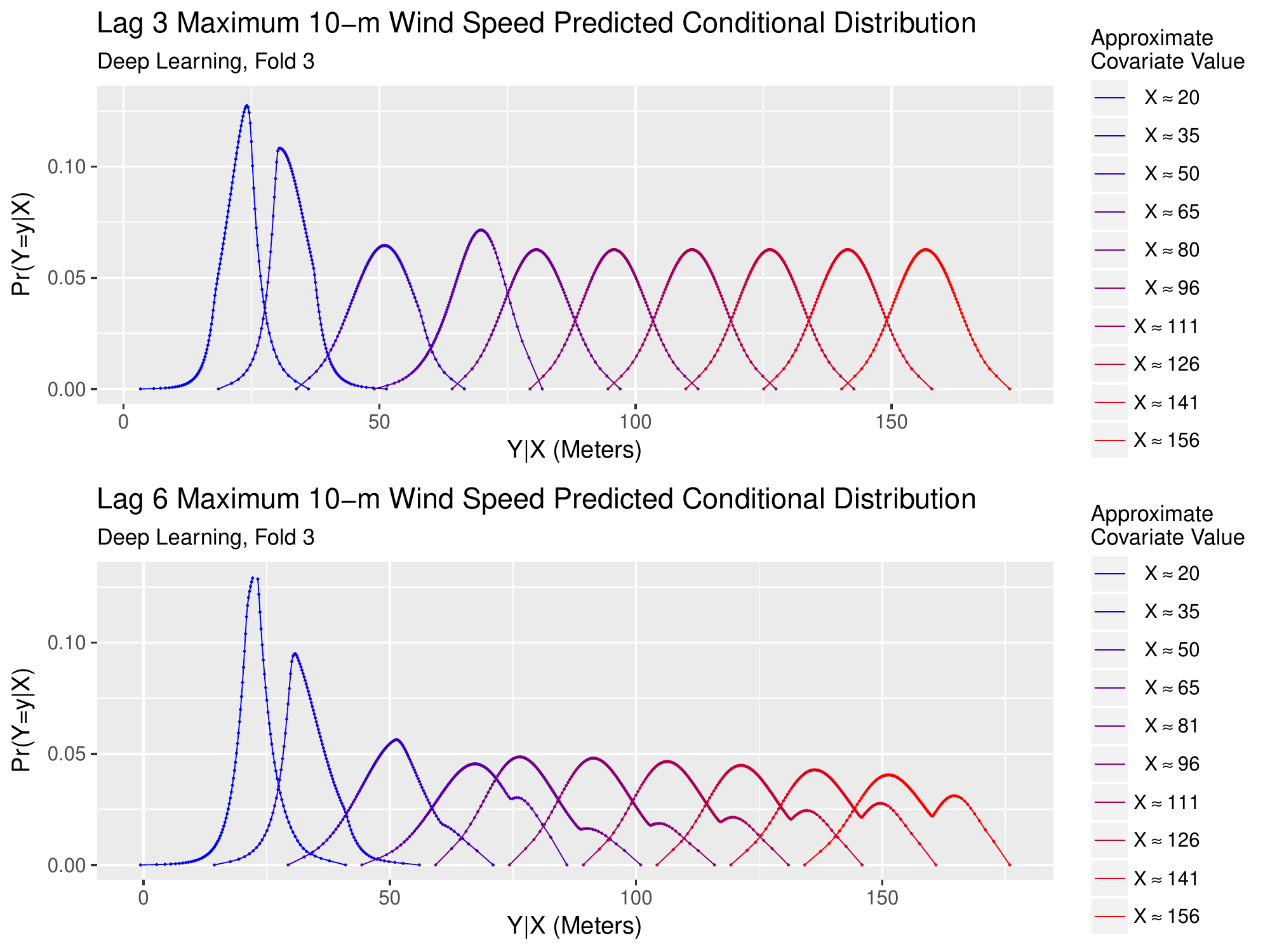}
		\caption{Deep learning IPP approximation ($\controls=20$) conditional maximum 10-meter wind speed distribution predictions for lag 3 and 6 using model constructed with observations from fold 3 as the testing dataset and all other observations as the training dataset. Estimated conditional response probabilities for 100 equally spaced quantiles sequenced between $0.5$ and $99.5$ are displayed with linear interpolation between quantiles. The $0.5$th and $99.5$th quantile density function values are rounded to 0. $\Pr(\Y=\y|\X)$ refers to the relative probability that the maximum 10-meter wind speed $\Y=\y$ occurs given the HWRF-forecasted maximum 10-meter wind speed value $\X$. $\Y|\X$ refers to the conditional response value $\Y$ given the covariate $\X$. \label{CDE_application_plot_fold3}}
\end{figure}

\begin{figure}[htbp!]
	\centering
		\includegraphics[width=1\textwidth]{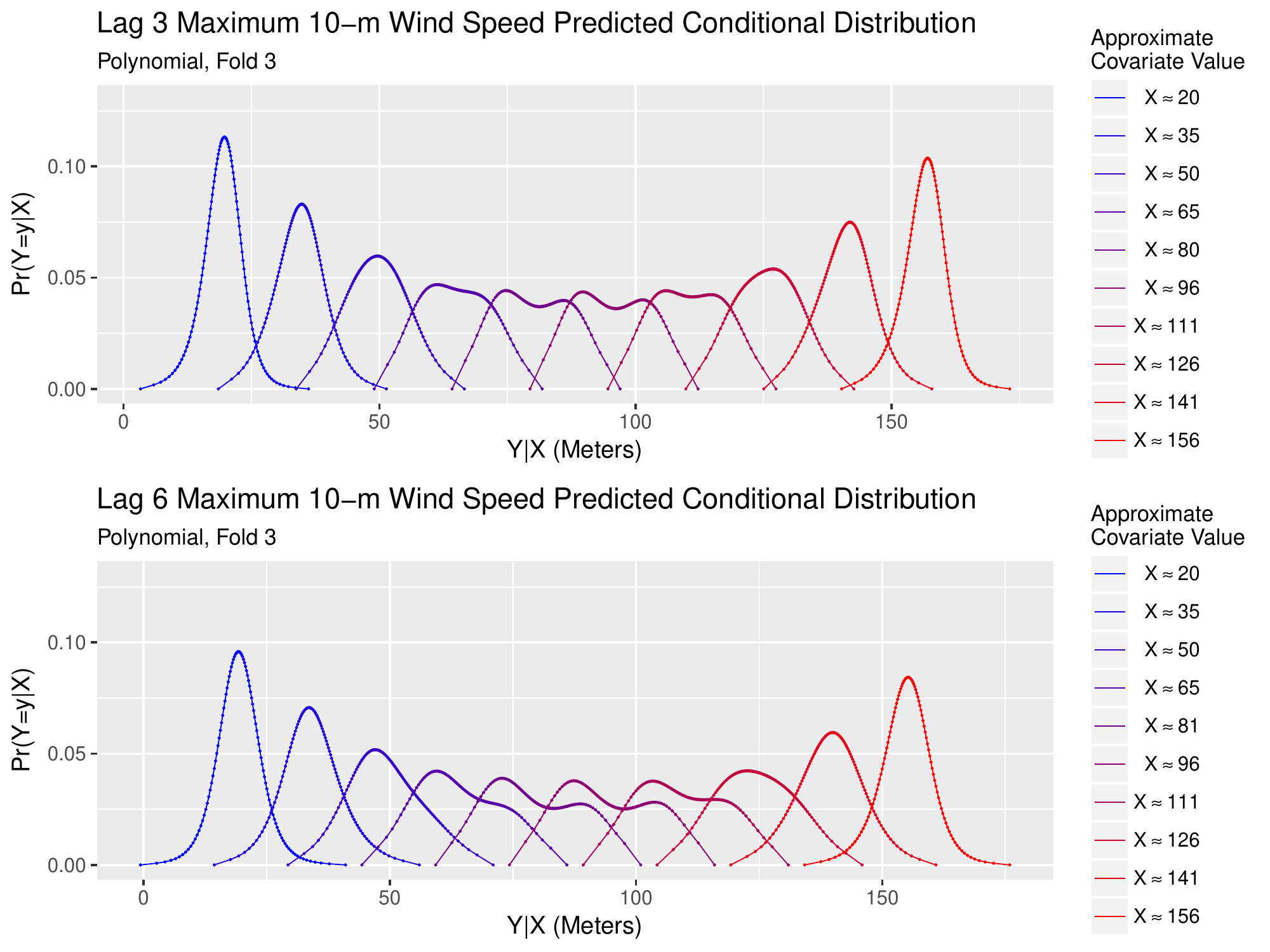}
		\caption{Polynomial MCC approximation ($\controls=1$) conditional maximum 10-meter wind speed distribution predictions for lag 3 and 6 using model constructed with with observations from fold 3 as the testing dataset and all other observations as the training dataset. Estimated conditional response probabilities for 100 equally spaced quantiles sequenced between $0.5$ and $99.5$ are displayed with linear interpolation between quantiles. The $0.5$th and $99.5$th quantile density function values are rounded to 0. $\Pr(\Y=\y|\X)$ refers to the relative probability that the maximum 10-meter wind speed $\Y=\y$ occurs given the HWRF-forecasted maximum 10-meter wind speed value $\X$. $\Y|\X$ refers to the conditional response value $\Y$ given the covariate $\X$. \label{Poly_CDE_application_plot_fold3}}
\end{figure}

\newpage

\end{appendices}

\renewcommand\refname{Literature Cited}
\bibliographystyle{chicago}
\bibliography{Proj2_Bibliography}

\end{document}